\documentclass{article} 
\usepackage{iclr2026_conference}
\usepackage{times}

\usepackage{amsmath,amsfonts,bm}

\def\eqref#1{equation~\ref{#1}}

\def\1{\bm{1}}

\def\vk{{\bm{k}}}

\def\vq{{\bm{q}}}

\def\vw{{\bm{w}}}

\def\vz{{\bm{z}}}

\def\mB{{\bm{B}}}

\def\mH{{\bm{H}}}

\def\mK{{\bm{K}}}

\def\mQ{{\bm{Q}}}

\def\mV{{\bm{V}}}

\DeclareMathAlphabet{\mathsfit}{\encodingdefault}{\sfdefault}{m}{sl}
\SetMathAlphabet{\mathsfit}{bold}{\encodingdefault}{\sfdefault}{bx}{n}
\newcommand{\tens}[1]{\bm{\mathsfit{#1}}}

\def\tX{{\tens{X}}}

\newcommand{\R}{\mathbb{R}}

\newcommand{\Z}{\mathbb{Z}}
\newcommand{\N}{\mathbb{N}}

\usepackage{hyperref}
\usepackage{url}
\usepackage{graphicx}
\usepackage{booktabs}
\usepackage{multirow}

\title{Structure-Aware Set Transformers: Temporal and Variable-Type Attention Biases for Asynchronous Clinical Time Series}

\author{
Joohyung Lee$^{1}$\ \quad
Kwanhyung Lee$^{1,2}$ \quad
Changhun Kim$^{1}$ \quad
Eunho Yang$^{1,2}$ \\
$^{1}$AITRICS \\
$^{2}$Korea Advanced Institute of Science and Technology (KAIST) \\
}

\iclrfinalcopy

\track{Research}

\begin{document}

\maketitle

\begin{abstract}
Electronic health records (EHR) are irregular, asynchronous multivariate time series. As time-series foundation models increasingly tokenize events rather than discretizing time, the input layout becomes a key design choice. Grids expose time$\times$variable structure but require imputation or missingness masks, risking error or sampling-policy shortcuts. Point-set tokenization avoids discretization but loses within-variable trajectories and time-local cross-variable context (Fig.~\ref{fig:fig1}). We restore these priors in \textbf{ST}ructure-\textbf{A}wa\textbf{R}e (\textbf{STAR}) \textbf{Set Transformer} by adding parameter-efficient soft attention biases: a temporal locality penalty $-|\Delta t|/\tau$ with learnable timescales and a variable-type affinity $B_{s_i,s_j}$ from a learned feature-compatibility matrix. We benchmark 10 depth-wise fusion schedules (Fig.~\ref{fig:fig2}). On three ICU prediction tasks, STAR-Set achieves AUC/APR of 0.7158/0.0026 (CPR), 0.9164/0.2033 (mortality), and 0.8373/0.1258 (vasopressor use), outperforming regular-grid, event-time grid, and prior set baselines. Learned $\tau$ and $B$ provide interpretable summaries of temporal context and variable interactions, offering a practical plug-in for context-informed time-series models.
\end{abstract}

\section{Introduction}
\label{sec:introduction}
Electronic health records (EHR) are irregularly sampled, asynchronous multivariate time series. Unlike images or text, there is no canonical discretization of time; consequently the \emph{input layout} presented to a neural encoder is a key design choice. Common layouts include (i) discretized \emph{regular grids}~\citep{lipton2015learning, labach2023duett, yu2024smart}, (ii) \emph{event-time} (sparse/irregular) grids ~\citep{che2018recurrent, chowdhury2023primenet}, and (iii) \emph{point-set} tokenizations over observed measurement events~\citep{horn2020set, tipirneni2022self}. Unless stated otherwise, we use \emph{representation} to denote this input layout.

In a \emph{regular-grid} layout, time is binned into fixed-width intervals (e.g., hourly) and measurements are aggregated per bin, yielding a dense tensor in $\mathbb{R}^{T\times V}$~\citep{lipton2015learning}. Because most variables are unobserved in most bins, grid models typically require a filling step, ranging from heuristic imputation (e.g., LOCF) to learned interpolation/imputation modules (e.g., mTAN)~\citep{lipton2015learning, shukla2021multi}. The \emph{event-time} grid instead indexes time by the union of observed timestamps $\tau$ and stores values in a sparse $\mathbb{R}^{|\tau|\times V}$ array; however, asynchrony still produces substantial missingness, motivating masks, time-gap features, or additional imputation~\citep{che2018recurrent}.

Attention-based sequence models~\citep{vaswani2017attention} further popularized event-time layouts because they can incorporate irregular timestamps via positional/continuous-time encodings and ignore padding through masking. Yet missingness is a double-edged signal in EHR: observation patterns reflect both physiology and care processes, so models can over-rely on missingness indicators and learn sampling-policy shortcuts that fail under domain shift (e.g., different measurement practices across hospitals)~\citep{che2018recurrent}.

\begin{figure*}[t]
  \centering
  \includegraphics[width=0.9\textwidth]{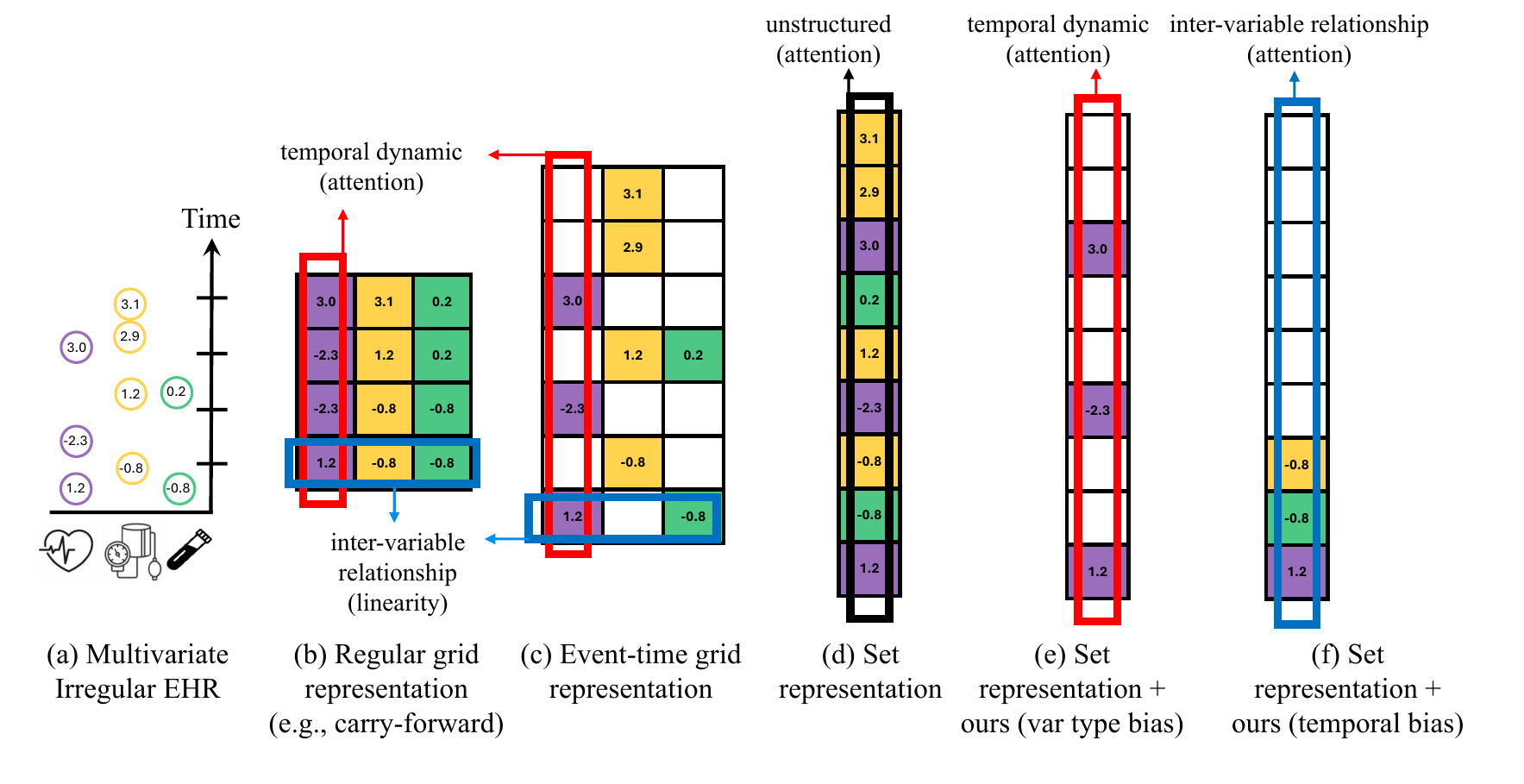}
  \caption{\textbf{EHR input layouts and biasing set attention.}
(a) Irregular, asynchronous EHR events. Grid and sparse time$\times$variable layouts (b,c) make within-variable trajectories (red) and time-local cross-variable relations (blue) explicit (sparse relies on missingness masks), whereas set tokenization (d) obscures both axes. We restore these inductive priors with a \emph{variable-type bias} (e), favoring same-variable interactions, and a \emph{temporal bias} (f), favoring temporally proximal interactions.}

  \label{fig:fig1}
\end{figure*}

Point-set representations avoid discretization by treating each observed event as a token (e.g., value, time, and variable identity) and applying permutation-invariant encoders~\citep{horn2020set}. This sidesteps populating unobserved grid entries and reduces dependence on heuristic imputation, but it also removes the two structural axes grids provide by construction: within-variable trajectories (``columns'') and contemporaneous cross-variable relations (``rows''). As illustrated in Figure~\ref{fig:fig1}, these inductive biases become implicit, forcing attention to recover them from data alone.

We address this gap by augmenting point-set EHR encoders with simple, parameter-efficient \emph{attention biases} that re-introduce grid-like structure while preserving set flexibility. We propose (i) a \emph{variable-type bias} that encourages interactions among tokens of the same measurement type and (ii) a \emph{temporal bias} that favors interactions among temporally nearby tokens (Figure~\ref{fig:fig1}). We then study where to inject these biases across Transformer depth and identify an effective layer-fusion strategy via systematic ablations over bias type and insertion schedule (Figure~\ref{fig:fig2}). More broadly, our approach echoes factorization choices in video modeling, where separating temporal dynamics from channel interactions improves efficiency~\citep{tran2018closer}.

\paragraph{Our contributions are:}
\begin{enumerate}
\item \textbf{Biasing set attention for irregular EHR:}
We introduce \textbf{STAR-Set Transformer} (vt--vt in Figure~\ref{fig:fig2}), which augments point-set EHR encoders with simple \emph{additive attention biases} to recover grid-like inductive structure without discretization.

\item \textbf{Two complementary, parameter-efficient biases:}
We combine (i) a \textbf{temporal bias} that favors temporally proximal interactions via a learnable time-distance penalty, and (ii) a \textbf{variable-type bias} parameterized by a learnable type-compatibility matrix.

\item \textbf{Systematic layer-wise ablations and consistent gains:}
We evaluate bias injection schedules across transformer depth and identify effective layer-fusion strategies, yielding consistent improvements on multiple ICU prediction tasks over grid and set-based baselines.

\end{enumerate}

\section{Method}
\label{sec:method}

\subsection{Set representation and its embedder for EHR}
We represent an EHR episode as an irregular, asynchronous \emph{set} of events~\citep{horn2020set}.
Given a mini-batch of size $B$, we pack episodes into a tensor
$\tX \in \R^{B \times L \times 3}$, where each event is a triplet
\begin{equation}
\tX_{b,i,:} = \big(t_{b,i},\, v_{b,i},\, s_{b,i}\big),
\end{equation}
with timestamp $t_{b,i}\in\R$, observed value $v_{b,i}\in\R$, and variable/type index
$s_{b,i}\in\{0,\dots,F-1\}$. Each episode has an effective length $\ell_b\le L$
(\texttt{input\_lengths}$\in\N^B$); indices $i>\ell_b$ are padding.
We use ITE~\citep{tipirneni2022self} as the set embedder and denote its output token sequence by
$\mH^{(0)}\in\R^{B\times S\times D}$ with $S=L+2$ (two special tokens plus $L$ event tokens).
The two special tokens are a learned \texttt{[CLS]} and a demographic token (\texttt{demo})
from a Linear--Tanh projection of age/sex.

\subsection{Transformer encoder with soft attention biases}
STAR-Set Transformer encodes $\mH^{(0)}$ with an $L_{\mathrm{enc}}$-layer (4 in this study)
pre-norm Transformer encoder ($H$ heads, GELU activations). Our core modification is to add
\emph{soft additive} biases to the attention logits, injecting (i) a temporal locality prior and
(ii) a variable-type prior that are explicit in grid/sparse layouts but not in set tokenizations
(Figure~\ref{fig:fig1}). See Appendix~\ref{app:per_token} for more details.

\begin{figure*}[t]
  \centering
  \includegraphics[width=0.9\textwidth]{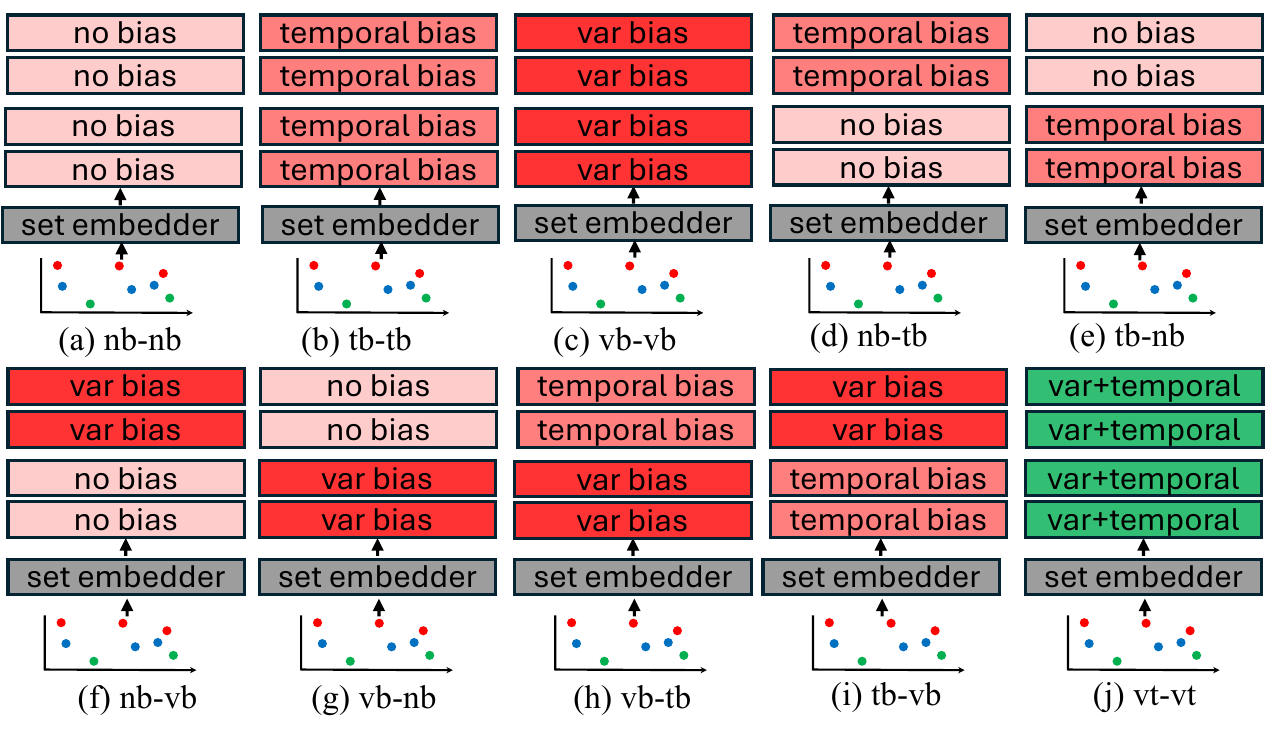}
  \caption{\textbf{Layer-wise fusion strategies for soft attention biases in the set encoder.}
Each panel illustrates a bias schedule applied across Transformer encoder layers (stacked blocks from early/lower to late/upper) on top of the set embedder.
We ablate \emph{no bias} (nb), \emph{temporal bias} (tb), \emph{variable-type bias} (vb), and their combination (vt).
The shorthand ``x--y'' denotes using bias x in the lower layers and bias y in the upper layers:
(a) nb--nb, (b) tb--tb, (c) vb--vb, (d) nb--tb, (e) tb--nb, (f) nb--vb, (g) vb--nb, (h) vb--tb, (i) tb--vb, and (j) vt--vt. We denote vt--vt as our proposed STAR Set Transformer.}

  \label{fig:fig2}
\end{figure*}

\paragraph{Temporal bias.}
For each layer $\ell$ and head $h$, we learn a timescale and define an additive time-distance
penalty between query token $i$ and key token $j$:
\begin{equation}
b^{(\ell,h)}_{\mathrm{time}}(b,i,j) \;=\; -\frac{\big|\tilde{t}_{b,i}-\tilde{t}_{b,j}\big|}{\tau_{\ell,h}},
\ \ \tau_{\ell,h} = \exp(\omega_{\ell,h}) + \varepsilon.
\label{eq:time_bias}
\end{equation}

\paragraph{Variable-type bias.}
For each layer $\ell$ and head $h$, we learn a type affinity matrix $\mB_{\ell,h}\in\R^{F\times F}$.
The variable-type bias is
\begin{equation}
b^{(\ell,h)}_{\mathrm{var}}(b,i,j) \;=\;
\begin{cases}
(\mB_{\ell,h})_{\tilde{s}_{b,i},\,\tilde{s}_{b,j}}, & \tilde{s}_{b,i},\tilde{s}_{b,j}\in\{0,\dots,F-1\},\\
0, & \text{otherwise}.
\end{cases}
\label{eq:var_bias}
\end{equation}

\paragraph{Biased attention logits.}
Let $d_h=D/H$, $\mQ^{(\ell,h)},\mK^{(\ell,h)},\mV^{(\ell,h)}$ be the standard projections, and
$m^{\mathrm{pad}}_{b,j}$ be the key padding mask (Appendix~\ref{app:setup}). We modify the attention
logits as
\begin{align}
A^{(\ell,h)}_{b,i,j} &=
\frac{\langle \vq^{(\ell,h)}_{b,i}, \vk^{(\ell,h)}_{b,j}\rangle}{\sqrt{d_h}}
\;+\; \1\!\left[\texttt{bias\_type}[\ell]\in\{\texttt{tb},\texttt{vtb}\}\right]\,
b^{(\ell,h)}_{\mathrm{time}}(b,i,j) \nonumber\\
&\quad+\; \1\!\left[\texttt{bias\_type}[\ell]\in\{\texttt{vb},\texttt{vtb}\}\right]\,
b^{(\ell,h)}_{\mathrm{var}}(b,i,j)
\;+\; m^{\mathrm{pad}}_{b,j}.
\label{eq:attn_logits}
\end{align}
We then apply the usual $\mathrm{softmax}$ over $j$ and compute the attention-weighted sum of values.

\paragraph{Layer-wise bias schedule.}
At each encoder layer $\ell$ we choose
\begin{equation}
\texttt{bias\_type}[\ell] \in \{\texttt{nb},\texttt{tb},\texttt{vb},\texttt{vtb}\},
\qquad \ell=1,\dots,L_{\mathrm{enc}},
\end{equation}
where \texttt{nb} applies no bias, \texttt{tb} applies temporal bias only, \texttt{vb} applies
variable-type bias only, and \texttt{vtb} applies both. In our ablations we often use two-stage
schedules of the form ``\texttt{a-b}'', meaning \texttt{a} in the lower layers and \texttt{b} in the
upper layers (Figure~\ref{fig:fig2}).

\paragraph{Prediction and supervised objective.}
After the final encoder layer and layer normalization, we use the \texttt{[CLS]} embedding as the
episode representation $\vz_b=\mH^{(L_{\mathrm{enc}})}_{b,0,:}\in\R^D$ and predict
$\hat{y}_b=\vw^\top\vz_b+c$. For binary outcomes $y_b\in\{0,1\}$, we train end-to-end with the
sigmoid cross-entropy loss (BCE with logits):
\begin{equation}
\mathcal{L}=\frac{1}{B}\sum_{b=1}^{B}\mathrm{BCEWithLogits}\!\left(\hat{y}_b,\;y_b\right).
\end{equation}

\begin{table*}[t]
\centering
\caption{Downstream prediction performance (AUC and APR) on three clinical tasks: CPR, mortality, and vasopressor use.}
\label{tab:main_results}
\small
\setlength{\tabcolsep}{3pt}
\renewcommand{\arraystretch}{1.15}
\begin{tabular}{lcc|cc|cc}
\hline
& \multicolumn{2}{c|}{\textbf{CPR}} & \multicolumn{2}{c|}{\textbf{Mortality}} & \multicolumn{2}{c}{\textbf{Vasopressor}} \\
\textbf{Model} & \textbf{AUC} & \textbf{APR} & \textbf{AUC} & \textbf{APR} & \textbf{AUC} & \textbf{APR} \\
\hline
\textbf{SMART (regular-grid) ~\citep{yu2024smart}} & 0.5098 & 0.0005 & 0.8457 & 0.0941 & 0.8239 & 0.0964 \\
\textbf{DueTT (regular-grid) ~\citep{labach2023duett}} & 0.6478 & 0.0010 & 0.8967 & 0.1527 & 0.8255 & 0.0929 \\
\textbf{PrimeNet (event-time grid) ~\citep{chowdhury2023primenet}} & 0.5492 & 0.0006 & 0.6289 & 0.0103 & 0.6956 & 0.025 \\
\textbf{STraTS (set) ~\citep{tipirneni2022self}} & 0.5397 & 0.0018 & 0.8778 & 0.1457 & 0.8241 & 0.1109 \\
\textbf{STAR Set Transformer (Ours)} & \textbf{0.7158} & \textbf{0.0026} & \textbf{0.9164} & \textbf{0.2033} & \textbf{0.8373} & \textbf{0.1258} \\
\hline
\end{tabular}
\end{table*}


\section{Experiments}
\label{sec:experiments}
\subsection{Experimental settings}
Please refer to Appendix ~\ref{app:setup} for the details about the dataset and implementation. Note that we set our experiments to figure out where and which attention priors should be injected within a framework of a 4-layer Transformer Encoder. Details are described in Section ~\ref{app:ablation_study}

\subsection{Results}
Table ~\ref{tab:main_results} compares STAR-Set Transformer to SMART and DueTT (regular-grid), PrimeNet (event-time grid), and STraTS (set) on CPR, Mortality, and Vasopressor prediction. STAR-Set Transformer achieves the best AUC and APR on all tasks: CPR reaches 0.7158 AUC (vs.\ 0.6478, DueTT) and 0.0026 APR (vs.\ 0.0018, STraTS); Mortality reaches 0.9164/0.2033 (vs.\ 0.8778/0.1457, STraTS); and Vasopressor reaches 0.8373 AUC (vs.\ 0.8255, DueTT) and 0.1258 APR (vs.\ 0.1109, STraTS). These gains suggest that our set-layout model effectively recovers grid-like inductive structure without committing to a fixed grid. Please refer to Appendix~\ref{app:ablation_study} for more detailed experimental results.

\section{Discussion and Conclusion}
Table~\ref{tab:main_results} shows that \textbf{STAR Set Transformer} achieves the best results on all reported tasks and metrics. It attains AUC/APR of 0.7158/0.0026 on CPR, 0.9164/0.2033 on Mortality, and 0.8373/0.1258 on Vasopressor. Relative to the strongest baselines, this corresponds to gains of +0.0680 AUC and +0.0008 APR on CPR, +0.0386 AUC and +0.0576 APR on Mortality, and +0.0118 AUC and +0.0149 APR on Vasopressor. Overall, STAR Set Transformer delivers consistent improvements over regular-grid, event-time-grid, and prior set-based approaches on these ICU prediction tasks.

\newpage

\bibliography{iclr2026_conference}
\bibliographystyle{iclr2026_conference}

\newpage

\appendix
\section{Dataset and Implementation Details}
\label{app:setup}

\subsection{Dataset and Cohort Construction}
We use MIMIC-IV ICU stays and follow the preprocessing protocol of \citet{lee2023learning}.
We evaluate three prediction tasks: CPR, mortality, and vasopressor need.
Unless otherwise stated, we split patients into train/validation/test sets with no patient overlap.
Models are selected by the best validation AUROC and evaluated once on the held-out test set.

\paragraph{Task definitions.}
For each task, we define an observation window and a prediction horizon following \citet{lee2023learning}.
Each prediction time requires at least 40 past observations within a 48-hour window and 20 future observations within a 12-hour window.

\begin{itemize}
  \item \textbf{CPR:} observation window = 48 hours (min 40 events), prediction horizon = 12 hours.
  \item \textbf{Vasopressor:} observation window = 48 hours (min 40 events), prediction horizon = 12 hours.
  \item \textbf{Mortality:} observation window = 48 hours (min 40 events), prediction horizon = 12 hours.
\end{itemize}

\paragraph{Cohort statistics.}
Table~\ref{tab:tab2} reports cohort sizes and class balance for each task.

\paragraph{Per-token metadata.}
\label{app:per_token}
Biases depend on each token's timestamp and variable ID. For event tokens we use their original
$(t_{b,i}, s_{b,i})$. For the two special tokens, we assign a reference time
\begin{equation}
t^{\mathrm{ref}}_b = \max_{1\le i \le \ell_b} t_{b,i},
\end{equation}
with fallback $t^{\mathrm{ref}}_b=0$ if $\ell_b=0$, and set their variable IDs to $-1$ so that
type bias is disabled. Collecting metadata over the full sequence ($S=L+2$),
\begin{equation}
\tilde{t}_{b}=\big[t^{\mathrm{ref}}_b,\; t^{\mathrm{ref}}_b,\; t_{b,1},\dots,t_{b,L}\big]\in\R^{S},
\qquad
\tilde{s}_{b}=\big[-1,\; -1,\; s_{b,1},\dots,s_{b,L}\big]\in\Z^{S}.
\end{equation}

\paragraph{Padding mask.}
Including the two special tokens, episode $b$ has $\ell_b+2$ valid tokens.
We use an \emph{additive} key padding mask $m^{\mathrm{pad}}_{b,j}$ with
$m^{\mathrm{pad}}_{b,j}=0$ for valid keys and $m^{\mathrm{pad}}_{b,j}=-\infty$ for padded keys.

\section{Ablation Study}
\label{app:ablation_study}


\paragraph{Goal.}
Figure~\ref{fig:fig2} evaluates \emph{where} and \emph{which} attention priors should be injected
when encoding point-set EHR tokens. In grid/event-time layouts, the time$\times$variable axes
implicitly encourage (i) within-variable temporal modeling and (ii) time-local cross-variable
context. In a set layout, these priors are not explicit; thus we ablate (a) the \emph{bias type}
(temporal vs.\ variable-type) and (b) the \emph{insertion depth} (early vs.\ late layers) while
keeping the backbone fixed.

\paragraph{Backbone held constant.}
All configurations share the same set embedder (ITE triplet embedding ~\citep{tipirneni2022self} with \texttt{[CLS]} and
\texttt{demo} and the same $L_{\mathrm{enc}}{=}4$-layer pre-norm Transformer encoder
(Section~\ref{sec:method}). The only change across ablations is the per-layer additive bias term in the
self-attention logits (Eq.~\ref{eq:attn_logits}).
For any layer assigned \texttt{nb}, the bias contribution is omitted (equivalently, the additive
mask is identically zero), so attention reduces to standard content-based dot-product attention
with padding masking.

\begin{table}[t]
\centering
\caption{Cohort statistics (MIMIC-IV).}
\label{tab:tab2}
\begin{tabular}{llcc}
\toprule
\textbf{Task} & \textbf{Split} & \textbf{\# Patients / Stays} & \textbf{\# Pos / \# Neg} \\
\midrule
\multirow{3}{*}{CPR}
 & Train & 29,544 / 37,156 & 561 / 1,578,395 \\
 & Val   & 6,341 / 8,160   & 54 / 31,345 \\
 & Test  & 6,292 / 7,934   & 151 / 326,508 \\
\midrule
\multirow{3}{*}{Vasopressor}
 & Train & 29,544 / 37,156 & 19,023 / 1,559,933 \\
 & Val   & 6,341 / 8,160   & 1,949 / 30,739 \\
 & Test  & 6,292 / 7,934   & 3,800 / 322,859 \\
\midrule
\multirow{3}{*}{Mortality}
 & Train & 29,544 / 37,156 & 7,705 / 1,571,251 \\
 & Val   & 6,341 / 8,160   & 751 / 31,304 \\
 & Test  & 6,292 / 7,934   & 1,706 / 324,953 \\
\bottomrule
\end{tabular}
\end{table}

\subsection{Bias families}
\label{app:bias_families}

We consider four mutually exclusive layer-level settings:
\begin{itemize}
  \item \texttt{nb}: \textbf{no bias} (baseline set encoder).
  \item \texttt{tb}: \textbf{temporal bias} only, using the linear time-distance penalty
  $b_{\mathrm{time}}^{(\ell,h)}(i,j)=-|\Delta t|/\tau_{\ell,h}$ (Eq.~\ref{eq:time_bias}).
  \item \texttt{vb}: \textbf{variable-type bias} only, using the learned type affinity
  $b_{\mathrm{var}}^{(\ell,h)}(i,j)=B_{\ell,h}[s_i,s_j]$ for observation--observation pairs
  (Eq.~\ref{eq:var_bias}).
  \item \texttt{vt}: \textbf{joint bias} (\texttt{vb}+\texttt{tb}), adding both terms to the logits.
\end{itemize}
Bias parameters are head- and layer-specific (learned $\tau_{\ell,h}$ and $B_{\ell,h}$), but the
\emph{schedule} below determines in which layers these parameters are active.

\begin{table}[t]
\centering
\caption{Layer-wise bias schedules for Fig.~\ref{fig:fig2} with $L_{\mathrm{enc}}{=}4$.
Each row specifies the bias applied at encoder layers 1--4 (early$\rightarrow$late).}
\label{tab:tab3}
\small
\setlength{\tabcolsep}{6pt}
\renewcommand{\arraystretch}{1.15}
\begin{tabular}{lcccc}
\hline
\textbf{Schedule} & \textbf{Layer 1} & \textbf{Layer 2} & \textbf{Layer 3} & \textbf{Layer 4} \\
\hline
\texttt{nb-nb} & \texttt{nb} & \texttt{nb} & \texttt{nb} & \texttt{nb} \\
\texttt{tb-tb} & \texttt{tb} & \texttt{tb} & \texttt{tb} & \texttt{tb} \\
\texttt{vb-vb} & \texttt{vb} & \texttt{vb} & \texttt{vb} & \texttt{vb} \\
\texttt{nb-tb} & \texttt{nb} & \texttt{nb} & \texttt{tb} & \texttt{tb} \\
\texttt{tb-nb} & \texttt{tb} & \texttt{tb} & \texttt{nb} & \texttt{nb} \\
\texttt{nb-vb} & \texttt{nb} & \texttt{nb} & \texttt{vb} & \texttt{vb} \\
\texttt{vb-nb} & \texttt{vb} & \texttt{vb} & \texttt{nb} & \texttt{nb} \\
\texttt{vb-tb} & \texttt{vb} & \texttt{vb} & \texttt{tb} & \texttt{tb} \\
\texttt{tb-vb} & \texttt{tb} & \texttt{tb} & \texttt{vb} & \texttt{vb} \\
\texttt{vt-vt} & \texttt{vt} & \texttt{vt} & \texttt{vt} & \texttt{vt} \\
\hline
\end{tabular}
\end{table}
\begin{table*}[t]
\centering
\caption{Performance (AUC and APR) across tasks for different layer-wise bias schedules.}
\small
\setlength{\tabcolsep}{4pt}
\begin{tabular}{lcccccccc}
\toprule
& \multicolumn{2}{c}{CPR} & \multicolumn{2}{c}{Mortality} & \multicolumn{2}{c}{Vasopressor} & \multicolumn{2}{c}{Average} \\
\cmidrule(lr){2-3}\cmidrule(lr){4-5}\cmidrule(lr){6-7}\cmidrule(lr){8-9}
Setting & AUC & APR & AUC & APR & AUC & APR & AUC & APR \\
\midrule
nb-nb   & 0.600 & 0.002 & 0.918 & 0.202 & 0.838 & 0.123 & 0.785 & 0.109 \\
tb-tb   & 0.753 & 0.003 & 0.908 & 0.182 & 0.836 & 0.118 & 0.832 & 0.101 \\
vb-vb   & 0.665 & 0.001 & 0.918 & 0.191 & 0.840 & 0.120 & 0.808 & 0.104 \\
nb-tb   & 0.668 & 0.005 & 0.914 & 0.189 & 0.834 & 0.124 & 0.805 & 0.106 \\
tb-nb   & 0.693 & 0.004 & 0.916 & 0.189 & 0.838 & 0.118 & 0.815 & 0.104 \\
nb-vb   & 0.657 & 0.003 & 0.918 & 0.199 & 0.834 & 0.121 & 0.803 & 0.107 \\
vb-nb   & 0.683 & 0.003 & 0.918 & 0.193 & 0.838 & 0.118 & 0.813 & 0.105 \\
vb-tb   & 0.699 & 0.006 & 0.919 & 0.184 & 0.836 & 0.116 & 0.818 & 0.102 \\
tb-vb   & 0.684 & 0.001 & 0.917 & 0.199 & 0.839 & 0.119 & 0.813 & 0.106 \\
vtb-vtb & 0.716 & 0.003 & 0.916 & 0.203 & 0.837 & 0.126 & 0.823 & 0.111 \\
\bottomrule
\end{tabular}
\label{tab:tab4}
\end{table*}

\subsection{Depth-wise fusion schedules}
\label{app:bias_schedules}

To isolate depth effects with minimal combinatorial growth, we use two-stage schedules of the form
\texttt{a-b}, where \texttt{a} is applied to the \emph{lower} half of layers and \texttt{b} to the
\emph{upper} half. With $L_{\mathrm{enc}}{=}4$, this corresponds to layers $\{1,2\}$ (early) and
$\{3,4\}$ (late). We evaluate the 10 schedules depicted in Fig.~\ref{fig:fig2}:

\paragraph{What these schedules test.}
The uniform settings (\texttt{tb-tb}, \texttt{vb-vb}, \texttt{vt-vt}) estimate the marginal value
of each inductive prior when applied throughout depth.
The swap schedules (\texttt{nb-tb} vs.\ \texttt{tb-nb} and \texttt{nb-vb} vs.\ \texttt{vb-nb})
test whether a given prior is more beneficial in early feature construction or late semantic
integration.
Finally, \texttt{vb-tb} and \texttt{tb-vb} test whether \emph{ordering} the priors (type-aware
mixing vs.\ temporal locality) across depth yields complementary benefits.

\subsection{Layer-wise bias schedules}
\label{sec:layerwise_bias}

To study \emph{where} to inject attention biases, we fix the encoder depth to $L_{\mathrm{enc}}{=}4$ and evaluate a set of two-stage schedules (Table~\ref{tab:tab3}). We denote no bias as \texttt{nb}, temporal bias as \texttt{tb}, variable-type bias as \texttt{vb}, and their combination as \texttt{vtb} (\texttt{vt} in the table). A schedule ``\texttt{a-b}'' applies bias \texttt{a} to early layers (1--2) and \texttt{b} to late layers (3--4), enabling controlled ablations of bias type versus injection depth.

Table~\ref{tab:tab4} reports downstream performance. Temporal bias is the main driver of AUC gains: \texttt{tb-tb} achieves the best average AUC (0.832), largely due to a large improvement on CPR (0.600$\rightarrow$0.753 AUC). Variable-type bias alone (\texttt{vb-vb}) yields smaller but consistent improvements over \texttt{nb-nb}. Combining both biases (\texttt{vtb-vtb}) provides the best average APR (0.111) while maintaining competitive AUC (0.823), and it attains the strongest APR on Mortality (0.203) and Vasopressor (0.126). Mixed schedules indicate that injecting biases only in part of the network is already beneficial, with a modest tendency for earlier-layer biasing (\texttt{tb-nb}, \texttt{vb-nb}) to outperform late-only variants (\texttt{nb-tb}, \texttt{nb-vb}).

\subsection{Experimental protocol and reporting}
\label{app:ablation_protocol}

For each schedule, we train the model under identical optimization, initialization, and data
processing settings; only the per-layer \texttt{bias\_type} list changes.
We report AUROC and average precision (APR) for each downstream task, and summarize performance
either (i) per task or (ii) by averaging across tasks after normalizing metrics (e.g., mean rank
or mean $\Delta$ relative to \texttt{nb-nb}).
To reduce variance, we recommend repeating each configuration with multiple random seeds and
reporting mean$\pm$std; all comparisons should use the same validation-based early stopping
criterion to avoid confounding depth schedules with training time.


\section{Limitations and Future Directions}
\label{app:limitations}

\paragraph{Limitations.}
\begin{itemize}
  \item \textbf{Limited hyperparameter sweep for bias parametrization.} We perform only a narrow search over (i) the initialization of the variable-type affinity matrix $\mB_{\ell,h}$ and (ii) the relative strength/scale of the temporal and variable-type bias terms (including $\tau$ initialization). Performance and stability may depend on these choices.
  \item \textbf{Single bias configuration per attention layer.} Our current design assigns one bias setting (\texttt{nb}/\texttt{tb}/\texttt{vb}/\texttt{vtb}) to an entire layer, so all heads share the same bias type. This may limit head specialization and prevent simultaneously modeling heterogeneous temporal and type-specific relations within a layer.
  \item \textbf{Supervised-only evaluation.} Experiments are restricted to supervised training on downstream ICU tasks. We do not yet evaluate STAR-Set under self-supervised pretraining, transfer/linear probing, or robustness settings (e.g., sampling-policy shift).
\end{itemize}

\paragraph{Future directions.}
\begin{itemize}
  \item \textbf{Systematic bias tuning and initialization study.} Conduct broader sweeps (or automated search) over $\mB$ initialization (e.g., identity vs.\ low-rank/random), bias scaling/regularization, and $\tau$ schedules to characterize sensitivity and identify robust defaults.
  \item \textbf{Head-wise (channel-wise) bias routing.} Generalize STAR-Set to allow different heads within a layer to use different bias types or learned mixtures (e.g., gated \texttt{tb}/\texttt{vb} per head), encouraging explicit specialization for temporal vs.\ type-driven interactions.
  \item \textbf{Beyond supervised learning.} Integrate biases into self-supervised objectives (e.g., masked event modeling) and evaluate transfer, label efficiency, and domain-shift robustness across cohorts/hospitals.
\end{itemize}

\end{document}